\documentclass[10pt,twocolumn]{IEEEtran}

\usepackage{cite}
\usepackage{graphicx}
\usepackage{epsfig}
\usepackage{psfrag}
\usepackage{subfigure}
\usepackage{url}
\usepackage{stfloats}
\usepackage{amsmath}
\usepackage{color}
\usepackage{amssymb}
\usepackage{epsfig}
\usepackage{epstopdf}

\usepackage{algorithm}
\usepackage{algorithmic}

\usepackage{array}

\begin{document}

\title{Extension of SBL Algorithms for the Recovery of Block Sparse Signals with Intra-Block Correlation}

\author{Zhilin Zhang$^*$, \IEEEmembership{Student Member, IEEE} and Bhaskar D. Rao, \IEEEmembership{Fellow, IEEE}
\thanks{Copyright (c) 2012 IEEE. Personal use of this material is permitted. However, permission to use this material for any other purposes must be obtained from the IEEE by sending a request to pubs-permissions@ieee.org.}
\thanks{Z.Zhang and B.D.Rao are with the Department of Electrical and Computer Engineering, University of California, San Diego, La Jolla, CA 92093-0407, USA. Email: zhangzlacademy@gmail.com (Z.Z), brao@ucsd.edu (B.D.R). The work was supported by NSF grants CCF-0830612 and CCF-1144258. \emph{Asterisk indicates corresponding author}.} }

\markboth{Published in IEEE Transactions on Signal Processing, vol. 61, no. 8, pp. 2009-2015, 2013}{Zhang \MakeLowercase{\textit{et al.}}}

\maketitle

\begin{abstract}
We examine the recovery of block sparse signals and extend the recovery framework in two important directions; one by exploiting the signals' intra-block correlation and the other by generalizing the signals' block structure. We propose two families of algorithms based on the framework of block sparse Bayesian learning (BSBL). One family, directly derived from the BSBL framework, require knowledge of the block structure. Another family, derived from an expanded BSBL framework, are based on a weaker assumption on the  block structure, and can be used when the block structure is completely unknown. Using these algorithms we show that exploiting intra-block correlation is very helpful in improving recovery performance. These algorithms also shed light on how to modify existing algorithms or design new ones to exploit such correlation and improve performance.
\end{abstract}


%

\IEEEpeerreviewmaketitle

\section{Introduction}
\label{sec:intro}
Sparse signal recovery and the associated problems of compressed sensing have received much attention in recent years \cite{elad2010sparse}. The basic model is given by
\begin{eqnarray}
\mathbf{y}= \mathbf{\Phi} \mathbf{x} + \mathbf{v},
\label{equ:SMV model}
\end{eqnarray}
where $\mathbf{y} \in \mathbb{R}^{M \times 1}$ is a known measurement vector, $\mathbf{\Phi} \in \mathbb{R}^{M \times N} (M \ll N)$ is a known matrix (generally called a basis matrix) and any $M$ columns are linearly independent, $\mathbf{x}\in \mathbb{R}^{N \times 1}$ is a sparse signal to be recovered, and $\mathbf{v}$ is an unknown noise vector. In applications,  $\mathbf{x}$ generally has additional structure. A widely studied structure is the block/group structure \cite{grouplasso,ModelCS,BOMP}. With this structure, $\mathbf{x}$ can be viewed as a concatenation of $g$ blocks, i.e.,
\begin{eqnarray}
\mathbf{x} = [ \underbrace{x_1,\cdots,x_{d_1}}_{\mathbf{x}_1^T},   \cdots,  \underbrace{x_{d_{g-1}+1},\cdots,x_{d_g}}_{\mathbf{x}_g^T}]^T
\label{equ:partition}
\end{eqnarray}
where $d_i (\forall i)$ are not necessarily identical. Among the $g$ blocks, only $k$ ($k \ll g$) blocks are nonzero but their locations are unknown. It is known that exploiting such block partition can further improve recovery performance.

A number of algorithms have been proposed to recover sparse signals with the block structure. Typical algorithms include Model-CoSaMp \cite{ModelCS}, Block-OMP \cite{BOMP}, and Group-Lasso type algorithms such as the original Group Lasso algorithm \cite{grouplasso}, Group Basis Pursuit \cite{van2008probing}, and Mixed $\ell_2/\ell_1$ Program \cite{Eldar2009}. These algorithms require knowledge of the block partition (\ref{equ:partition}). Other algorithms, such as StructOMP \cite{Huang2009}, do not need to know the block partition but need to know other \emph{a priori} information, e.g., the number of nonzero elements in $\mathbf{x}$. Recently, CluSS-MCMC \cite{bcs_mcmc} and BM-MAP-OMP \cite{faktor2010exploiting} have been proposed, which require very little \emph{a priori} knowledge.

However, few existing algorithms consider intra-block correlation, i.e., the amplitude correlation among the elements within each block. In practical applications intra-block correlation widely exists in signals, such as physiological signals \cite{zhang_ECG} and images. In this work we derive several algorithms that explore and exploit intra-block correlation to improve performance. These algorithms are based on our recently proposed block sparse Bayesian learning (BSBL) framework \cite{Zhilin2011c}. Although the framework was initially used to derive algorithms for a multiple measurement vector (MMV) model \cite{Cotter2005}, it has not been used for the block sparse model (\ref{equ:SMV model})-(\ref{equ:partition}). The successes of sparse Bayesian learning methods in past contexts motivate us to consider their extension to this problem and fill this gap.

One contribution of our work is that the proposed algorithms are the first ones in the category that adaptively explore and exploit intra-block correlation. Experiments showed that the  developed algorithms significantly outperform competitive algorithms. We also suggest a promising strategy to incorporate intra-block correlation in the Group-Lasso type algorithms to improve their performance.

Another contribution is the finding of the effect of  intra-block correlation on algorithms' performance. We find the effect of intra-block correlation is quite different from the effect of temporal correlation in an MMV model \cite{Zhilin2011c}. This is interesting, since an MMV model can be viewed as a special case of a block sparse model and temporal correlation in an MMV model corresponds to  intra-block correlation in a block sparse model.

The third contribution is the development of a simple approximate model and corresponding algorithms to solve the problem when the block partition is entirely unknown. These algorithms are effective especially in noisy environments.

In this paper bold symbols are reserved for vectors and matrices. For square matrices $\mathbf{A}_1,\cdots,\mathbf{A}_g$, $\mathrm{diag}\{\mathbf{A}_1,\cdots,\mathbf{A}_g \}$ denotes a block diagonal matrix with principal diagonal blocks being $\mathbf{A}_1,\cdots,\mathbf{A}_g$ in turn. $\mathrm{Tr}(\mathbf{A})$ denotes the trace of $\mathbf{A}$. $\boldsymbol{\gamma} \succeq \mathbf{0} $ means each element in the vector $\boldsymbol{\gamma}$ is nonnegative.

Parts of this work have been published in \cite{Zhilin_ICASSP2012}.

\section{Overview of the BSBL Framework}
\label{sec:overview}

This section briefly describes the BSBL framework \cite{Zhilin2011c}, upon which we develop our algorithms. In this framework, each block $\mathbf{x}_i \in \mathbb{R}^{d_i \times 1}$ is assumed to satisfy a parameterized multivariate Gaussian distribution:
\begin{eqnarray}
p(\mathbf{x}_i; \gamma_i, \mathbf{B}_i) \sim  \mathcal{N}(\textbf{0},\gamma_i \mathbf{B}_i), \quad i=1,\cdots,g
\nonumber
\end{eqnarray}
with the unknown parameters $\gamma_i$ and $\mathbf{B}_i$. Here $\gamma_i$ is a nonnegative parameter controlling the block-sparsity of $\mathbf{x}$. When $\gamma_i=0$, the $i$-th block becomes zero. During the learning procedure most $\gamma_i $ tend to be zero, due to the mechanism of automatic relevance determination \cite{Tipping2001}. Thus sparsity at the block level is encouraged. $\mathbf{B}_i \in \mathbb{R}^{d_i \times d_i}$ is a positive definite matrix, capturing the correlation structure of the $i$-th block. Under the assumption that blocks are mutually uncorrelated, the prior of $\mathbf{x}$ is $p(\mathbf{x};\{\gamma_i,\mathbf{B}_i\}_i) \sim  \mathcal{N}(\textbf{0},\mathbf{\Sigma}_0)$, where $\mathbf{\Sigma}_0 = \mathrm{diag}\{\gamma_1 \mathbf{B}_1,\cdots,\gamma_g \mathbf{B}_g\}$. Assume the noise vector satisfies $p(\mathbf{v};\lambda) \sim  \mathcal{N}(\textbf{0},\lambda \mathbf{I})$, where $\lambda$ is a positive scalar. Therefore the posterior of $\mathbf{x}$ is given by
\begin{eqnarray}
p(\mathbf{x}|\mathbf{y}; \lambda, \{\gamma_i, \mathbf{B}_i\}_{i=1}^g) = \mathcal{N}(\boldsymbol{\mu}_x, \mathbf{\Sigma}_x) \nonumber
\end{eqnarray}
with
\begin{eqnarray}
\boldsymbol{\mu}_x  &=& \mathbf{\Sigma}_0 \mathbf{\Phi}^T \big(\lambda \mathbf{I} +  \mathbf{\Phi} \mathbf{\Sigma}_0 \mathbf{\Phi}^T \big)^{-1} \mathbf{y}, \nonumber \\
\mathbf{\Sigma}_x &=& (\mathbf{\Sigma}_0^{-1} + \frac{1}{\lambda}\mathbf{\Phi}^T \mathbf{\Phi})^{-1}. \nonumber
\end{eqnarray}
Once the parameters $\lambda, \{\gamma_i, \mathbf{B}_i\}_{i=1}^g$ are estimated, the Maximum-A-Posteriori (MAP) estimate of $\mathbf{x}$, denoted by $\widehat{\mathbf{x}}$, can be directly obtained from the mean of the posterior, i.e.,
\begin{eqnarray}
\widehat{\mathbf{x}} \leftarrow  \mathbf{\Sigma}_0 \mathbf{\Phi}^T \big(\lambda \mathbf{I} +  \mathbf{\Phi} \mathbf{\Sigma}_0 \mathbf{\Phi}^T \big)^{-1} \mathbf{y}. \nonumber
\end{eqnarray}

The parameters can be estimated by a Type II maximum likelihood procedure \cite{Tipping2001}. This is equivalent to minimizing the following cost function
\begin{eqnarray}
\mathcal{L}(\Theta) &\triangleq & -2 \log \int p(\mathbf{y}|\mathbf{x};\lambda) p(\mathbf{x};\{\gamma_i,\mathbf{B}_i\}_i) d \mathbf{x}  \nonumber \\
 &=& \log|\lambda \mathbf{I} + \mathbf{\Phi} \mathbf{\Sigma}_0 \mathbf{\Phi}^T  |  + \mathbf{y}^T (\lambda \mathbf{I} + \mathbf{\Phi} \mathbf{\Sigma}_0 \mathbf{\Phi}^T)^{-1} \mathbf{y}, \nonumber \\
\label{equ:costfunc}
\end{eqnarray}
where $\Theta$ denotes all the parameters, i.e., $\Theta \triangleq \{\lambda,\{\gamma_i,\mathbf{B}_i\}_{i=1}^g \}$. This framework is called the BSBL framework \cite{Zhilin2011c}.

Each algorithm derived from this framework includes three learning rules, i.e., the learning rules for $\gamma_i$, $\mathbf{B}_i$, and $\lambda$. The learning rule for $\gamma_i$ is the main body of an algorithm. Different $\gamma_i$ learning rules lead to different convergence speed \footnote{The $\lambda$ learning rule also affects the speed, but its effect is not dominant.}, and determine the  best possible recovery performance when optimal values of $\lambda$ and $\mathbf{B}_i$ are given.

The $\lambda$ learning rule is important as well. If an optimal (or a good sub-optimal) value for $\lambda$ cannot be obtained, the recovery performance can be very poor even if the $\gamma_i$ learning rule could potentially lead to perfect recovery performance.

As for $\mathbf{B}_i (\forall i)$, it can be shown \cite{Zhilin2011c} that in noiseless environments, the global minimum of (\ref{equ:costfunc}) always leads to the true sparse solution irrespective of the value of $\mathbf{B}_i$; $\mathbf{B}_i$ only affects local convergence (such as changing the shape of the basins of attraction of local minima). Therefore, one can impose various constraints on the form of $\mathbf{B}_i$ to achieve better performance and  prevent overfitting.

An interesting property of the framework is that it is capable of directly recovering less-sparse or non-sparse signals as shown in \cite{zhang_ECG}.

\section{Algorithms When the Block Partition is Known}
\label{sec:type1}

In this section we propose three algorithms, which require knowledge of the block partition (\ref{equ:partition}).

\subsection{BSBL-EM: the Expectation-Maximization Method}

This algorithm can be readily derived from our previous work \cite{Zhilin2011c} on an MMV model with suitable adaptation. Thus we omit details on algorithm derivation. However, several necessary changes, particularly for enhancing the robustness of the learning rules for $\lambda$ and $\mathbf{B}_i$, have to be made here.

Following the Expectation Maximization (EM) method \cite{Zhilin2011c}, we can derive the learning rules for $\gamma_i$ and $\lambda$:
\begin{eqnarray}
\gamma_i  &\leftarrow &  \frac{1}{d_i} \mathrm{Tr}\big[ \mathbf{B}_i^{-1} \big( \mathbf{\Sigma}_x^i + \boldsymbol{\mu}_x^i (\boldsymbol{\mu}_x^i)^T\big)\big],\quad \forall i \label{equ:EM_gammaRule} \\
\lambda  & \leftarrow & \frac{\|\mathbf{y}-\mathbf{\Phi} \boldsymbol{\mu}_x\|_2^2 + \mathrm{Tr}( \mathbf{\Sigma}_x \mathbf{\Phi}^T \mathbf{\Phi})}{M},
\label{equ:EM_lambdaRule}
\end{eqnarray}
where  $\boldsymbol{\mu}_x^i \in \mathbb{R}^{d_i \times 1}$ is the corresponding $i$-th block in $\boldsymbol{\mu}_x$, and $\mathbf{\Sigma}_x^{i} \in \mathbb{R}^{d_i \times d_i}$ is the corresponding $i$-th principal diagonal block in $\mathbf{\Sigma}_x$. Note that the $\lambda$ learning rule (\ref{equ:EM_lambdaRule}) is not robust in low SNR cases.
By numerical study, we empirically find that this is due in part to the disturbance caused by the off-block-diagonal elements in $\mathbf{\Sigma}_x$ and $\mathbf{\Phi}^T \mathbf{\Phi}$. Therefore, we set their off-block-diagonal elements to zero, leading to the learning rule
\begin{eqnarray}
\lambda  &\leftarrow& \frac{\|\mathbf{y}-\mathbf{\Phi} \boldsymbol{\mu}_x\|_2^2 + \sum_{i=1}^g \mathrm{Tr}( \mathbf{\Sigma}_x^i (\mathbf{\Phi}^i)^T \mathbf{\Phi}^i)}{M},
\label{equ:EM_lambdaRule_robust}
\end{eqnarray}
where $\mathbf{\Phi}^i \in \mathbb{R}^{M \times d_i}$ is the submatrix of $\mathbf{\Phi}$, which corresponds to the $i$-th block of $\mathbf{x}$. This $\lambda$ learning rule is better than (\ref{equ:EM_lambdaRule}) in generally noisy environments (e.g., $\mathrm{SNR} < 20 \mathrm{dB}$). In noiseless cases there is no need to use any $\lambda$ learning rules. Just fixing $\lambda$ to a  small value, e.g., $10^{-10}$, can yield satisfactory performance.

Similar to \cite{Zhilin2011c}, using the EM method we can derive a learning rule for $\mathbf{B}_i$. However, assigning a different $\mathbf{B}_i$ to each block can result in overfitting. When blocks have the same size, an effective strategy to avoid the overfitting is  parameter averaging \cite{Zhilin2011c}, i.e., constraining $\mathbf{B}_i = \mathbf{B} (\forall i)$. Using this constraint, the learning rule for $\mathbf{B}$ can be derived as follows
\begin{eqnarray}
\mathbf{B} \leftarrow \frac{1}{g} \sum_{i=1}^g \frac{\mathbf{\Sigma}_x^i + \boldsymbol{\mu}_x^i (\boldsymbol{\mu}_x^i)^T}{\gamma_i}.
\label{equ:EM_Brule}
\end{eqnarray}
However, the algorithm's performance can be improved by further constraining the matrix $\mathbf{B}$. The idea is to find a positive definite and symmetric matrix $\widehat{\mathbf{B}}$ such that it is determined by one parameter but is close to $\mathbf{B}$ especially along the main diagonal and the main sub-diagonal. Further, we find that for many applications modeling elements of a block as a first-order Auto-Regressive (AR) process is sufficient to model  intra-block correlation. In this case, the corresponding correlation matrix of the block is a Toeplitz matrix with the following form:
\begin{eqnarray}
\mathrm{Toeplitz}([1,r,\cdots,r^{d-1}])
= \left[
\begin {array}{cccc}
1             & r                     & \cdots   & r^{d-1} \\
\vdots        &         &                 & \vdots     \\
r^{d-1}       & r^{d-2}       &  \cdots               & 1
\end {array}
\right] \label{equ:EM_Brule_samesize}
\end{eqnarray}
where $r$ is the AR coefficient and $d$ is the block size. Here we constrain $\widehat{\mathbf{B}}$ to have this form. Instead of estimating $r$ from the BSBL cost function, we empirically calculate its value by $r \triangleq \frac{m_1}{m_0}$, where $m_0$ (resp. $m_1$) is the average of the elements along the main diagonal (resp. the main sub-diagonal) of the matrix $\mathbf{B}$ in (\ref{equ:EM_Brule}).

When blocks have different sizes, the above idea can still be used. First, using the EM method we can derive the rule for each $\mathbf{B}_i$: $ \mathbf{B}_i \leftarrow \frac{1}{\gamma_i} \big[ \mathbf{\Sigma}_x^i + \boldsymbol{\mu}_x^i (\boldsymbol{\mu}_x^i)^T \big]$.
Then, for each $\mathbf{B}_i$ we calculate the averages of the elements along the main diagonal and the main sub-diagonal, i.e., $m_0^i$ and $m_1^i$, respectively, and average $m_0^i$ and $m_1^i$ for all blocks as follows: $\overline{m}_0 \triangleq \sum_{i=1}^g m_0^i$ and $\overline{m}_1 \triangleq \sum_{i=1}^g m_1^i$. Finally, we have $\overline{r} \triangleq \frac{\overline{m}_1}{\overline{m}_0}$, from which we construct   $\widehat{\mathbf{B}}_i$ for the $i$-th block:
\begin{eqnarray}
\widehat{\mathbf{B}}_i=\mathrm{Toeplitz}([1,\overline{r},\cdots,\overline{r}^{d_i-1}])  \quad (\forall i)
\label{equ:EM_Brule_diffsize}
\end{eqnarray}
We denote the above algorithm by \textbf{BSBL-EM}.

\subsection{BSBL-BO: the Bound-Optimization Method}

The BSBL-EM algorithm has satisfactory recovery performance but is slow. This is mainly due to the EM-based $\gamma_i$ learning rule. For the basic SBL algorithm, Tipping \cite{Tipping2001} derived a fixed-point based $\gamma_i$ learning rule to replace the EM-based one, which has faster convergence speed but is not robust in some noisy environments. Here we derive a fast $\gamma_i$ learning rule based on the bound-optimization method (also known as the Majorization-Minimization method) \cite{elad2010sparse,stoica2011spice}. The algorithm adopting this $\gamma_i$ learning rule is denoted by \textbf{BSBL-BO} (it uses the same learning rules for $\mathbf{B}_i$ and $\lambda$ as  BSBL-EM). It not only has fast speed, but also has satisfactory performance.

Note that the original cost function (\ref{equ:costfunc}) consists of two terms. The first term $ \log|\lambda \mathbf{I} + \mathbf{\Phi}\mathbf{\Sigma}_0 \mathbf{\Phi}^T|$ is concave with respect to $\boldsymbol{\gamma} \succeq \mathbf{0} $, where  $\boldsymbol{\gamma} \triangleq [\gamma_1,\cdots,\gamma_g]^T$. The second term $\mathbf{y}^T (\lambda \mathbf{I} + \mathbf{\Phi}\mathbf{\Sigma}_0 \mathbf{\Phi}^T)^{-1} \mathbf{y}$ is convex with respect to $\boldsymbol{\gamma} \succeq \mathbf{0} $. Since our goal is to minimize the cost function, we choose to find an upper-bound for the first item and then minimize the upper-bound of the cost function.

We use the supporting hyperplane of the first term as its upper-bound. Let $\boldsymbol{\gamma}^*$ be a given point in the $\boldsymbol{\gamma}$-space. We have
\begin{eqnarray}
\log|\lambda \mathbf{I} + \mathbf{\Phi}\mathbf{\Sigma}_0 \mathbf{\Phi}^T| &\leq & \log |\lambda \mathbf{I} + \mathbf{\Phi} \mathbf{\Sigma}_0^* \mathbf{\Phi}^T| \nonumber \\
 && + \sum_{i=1}^g \mathrm{Tr}\big( (\mathbf{\Sigma}_y^*)^{-1} \mathbf{\Phi}^i \mathbf{B}_i (\mathbf{\Phi}^i)^T) (\gamma_i - \gamma_i^*) \nonumber \\
& = & \sum_{i=1}^g \mathrm{Tr}\big( (\mathbf{\Sigma}_y^*)^{-1} \mathbf{\Phi}^i \mathbf{B}_i (\mathbf{\Phi}^i)^T) \gamma_i \nonumber \\
&& + \log |\mathbf{\Sigma}_y^*| \nonumber\\
 &&- \sum_{i=1}^g \mathrm{Tr}\big( (\mathbf{\Sigma}_y^*)^{-1} \mathbf{\Phi}^i \mathbf{B}_i (\mathbf{\Phi}^i)^T) \gamma_i^* \label{equ:MM}
\end{eqnarray}
where $\mathbf{\Sigma}_y^* = \lambda \mathbf{I} + \mathbf{\Phi}\mathbf{\Sigma}_0^* \mathbf{\Phi}^T$ and $\mathbf{\Sigma}_0^* \triangleq \mathbf{\Sigma}_0 |_{\boldsymbol{\gamma} = \boldsymbol{\gamma}^*}$.  Substituting (\ref{equ:MM}) into the cost function (\ref{equ:costfunc}) we have
\begin{eqnarray}
\mathcal{L}(\boldsymbol{\gamma})
& \leq & \sum_{i=1}^g \mathrm{Tr}\big( (\mathbf{\Sigma}_y^*)^{-1} \mathbf{\Phi}^i \mathbf{B}_i (\mathbf{\Phi}^i)^T) \gamma_i \nonumber\\
 && + \mathbf{y}^T (\lambda \mathbf{I} + \mathbf{\Phi} \mathbf{\Sigma}_0 \mathbf{\Phi}^T)^{-1} \mathbf{y}  + \log |\mathbf{\Sigma}_y^*| \nonumber \\
 && - \sum_{i=1}^g \mathrm{Tr}\big( (\mathbf{\Sigma}_y^*)^{-1} \mathbf{\Phi}^i \mathbf{B}_i (\mathbf{\Phi}^i)^T) \gamma_i^* \nonumber \\
&\triangleq & \widetilde{\mathcal{L}}(\boldsymbol{\gamma}) \label{equ:costfunc_tild}
\end{eqnarray}
The function $\widetilde{\mathcal{L}}(\boldsymbol{\gamma})$ is convex over $\boldsymbol{\gamma}$, and when $\boldsymbol{\gamma} = \boldsymbol{\gamma}^*$ we have $\mathcal{L}(\boldsymbol{\gamma}^*) = \widetilde{\mathcal{L}}(\boldsymbol{\gamma}^*)$. Further, for any $\boldsymbol{\gamma}_{\mathrm{min}}$ which minimizes $\widetilde{\mathcal{L}}(\boldsymbol{\gamma})$, we have the following relationship: $\mathcal{L}(\boldsymbol{\gamma}_{\mathrm{min}}) \leq \widetilde{\mathcal{L}}(\boldsymbol{\gamma}_{\mathrm{min}}) \leq \widetilde{\mathcal{L}}(\boldsymbol{\gamma}^*) = \mathcal{L}(\boldsymbol{\gamma}^*)$.
This indicates that when we minimize the surrogate function $\widetilde{\mathcal{L}}(\boldsymbol{\gamma})$ over $\boldsymbol{\gamma}$, the resulting minimum point effectively decreases the original cost function $\mathcal{L}(\boldsymbol{\gamma})$. We can use any optimization software to optimize (\ref{equ:costfunc_tild}). However, our experiments showed that this could take more time than BSBL-EM and lead to poorer recovery performance.
Therefore, we consider another surrogate function.

Using the identity
\begin{eqnarray}
\mathbf{y}^T (\lambda \mathbf{I} + \mathbf{\Phi} \mathbf{\Sigma}_0 \mathbf{\Phi}^T)^{-1} \mathbf{y} \equiv \min_\mathbf{x} \big[\frac{1}{\lambda} \|\mathbf{y}-\mathbf{\Phi x}\|_2^2 + \mathbf{x}^T \mathbf{\Sigma}_0^{-1} \mathbf{x}  \big],
\label{equ:identityEqu}
\end{eqnarray}
where the optimal $\mathbf{x}$ is $\boldsymbol{\mu}_x$, we have
\begin{eqnarray}
\widetilde{\mathcal{L}}(\boldsymbol{\gamma}) &=& \min_\mathbf{x} \frac{1}{\lambda} \|\mathbf{y}-\mathbf{\Phi x}\|_2^2 + \mathbf{x}^T \mathbf{\Sigma}_0^{-1} \mathbf{x}  \nonumber \\
&& + \sum_{i=1}^g \mathrm{Tr}\big( (\mathbf{\Sigma}_y^*)^{-1} \mathbf{\Phi}^i \mathbf{B}_i (\mathbf{\Phi}^i)^T) \gamma_i + \log |\mathbf{\Sigma}_y^*| \nonumber \\
  && - \sum_{i=1}^g \mathrm{Tr}\big( (\mathbf{\Sigma}_y^*)^{-1} \mathbf{\Phi}^i \mathbf{B}_i (\mathbf{\Phi}^i)^T) \gamma_i^*. \nonumber
\end{eqnarray}
Then, a new function
\begin{eqnarray}
\mathcal{G}(\boldsymbol{\gamma},\mathbf{x}) &\triangleq  & \frac{1}{\lambda} \|\mathbf{y}-\mathbf{\Phi x}\|_2^2 + \mathbf{x}^T \mathbf{\Sigma}_0^{-1} \mathbf{x}  \nonumber \\
&& + \sum_{i=1}^g \mathrm{Tr}\big( (\mathbf{\Sigma}_y^*)^{-1} \mathbf{\Phi}^i \mathbf{B}_i (\mathbf{\Phi}^i)^T) \gamma_i \nonumber \\
 && + \log |\mathbf{\Sigma}_y^*| - \sum_{i=1}^g \mathrm{Tr}\big( (\mathbf{\Sigma}_y^*)^{-1} \mathbf{\Phi}^i \mathbf{B}_i (\mathbf{\Phi}^i)^T) \gamma_i^* \nonumber
\end{eqnarray}
is defined, which is the upper-bound of $\widetilde{\mathcal{L}}(\boldsymbol{\gamma})$. Note that $\mathcal{G}(\boldsymbol{\gamma},\mathbf{x})$ is convex in both $\boldsymbol{\gamma}$ and $\mathbf{x}$. It can be easily shown that the solution $(\boldsymbol{\gamma}^\diamond)$ of $\widetilde{\mathcal{L}}(\boldsymbol{\gamma})$ is the solution $(\boldsymbol{\gamma}^\diamond, \mathbf{x}^\diamond)$ of $\mathcal{G}(\boldsymbol{\gamma},\mathbf{x})$. Thus, $\mathcal{G}(\boldsymbol{\gamma},\mathbf{x})$ is our final surrogate cost function.

Taking the derivative of $\mathcal{G}$ with respect to $\gamma_i$, we can obtain
\begin{eqnarray}
\gamma_i & \leftarrow & \sqrt{\frac{\mathbf{x}_i^T \mathbf{B}_i^{-1} \mathbf{x}_i}{\mathrm{Tr}\big( (\mathbf{\Phi}^i)^T (\mathbf{\Sigma}_y^*)^{-1} \mathbf{\Phi}^i \mathbf{B}_i )}}.  \label{equ:CV_gamma}
\end{eqnarray}
Due to this $\gamma_i$ learning rule, BSBL-BO  requires far fewer iterations than BSBL-EM, but both algorithms have comparable performance.

\subsection{BSBL-$\ell_1$: Hybrid of BSBL and Group-Lasso Type Algorithms}
Essentially, BSBL-EM and BSBL-BO operate in the $\boldsymbol{\gamma}$-space, since their cost function is a function of $\boldsymbol{\gamma}$. In contrast, most existing algorithms for the block sparse model (\ref{equ:SMV model})-(\ref{equ:partition}) directly operate in the $\mathbf{x}$-space, minimizing a data fit term and a penalty, which are both functions of $\mathbf{x}$. It is interesting to see  the relation between our BSBL algorithms and those algorithms.

Using the idea we presented in \cite{Zhilin2011b}, an extension of the duality space analysis for the basic SBL framework \cite{David2010},  we can transform the BSBL cost function (\ref{equ:costfunc}) from the $\boldsymbol{\gamma}$-space to the $\mathbf{x}$-space. Since $\lambda$ and $\mathbf{B}_i(\forall i)$ can be viewed as regularizers, for convenience we first treat them as fixed values.

First, using the identity (\ref{equ:identityEqu}) we can upper-bound the BSBL cost function as follows:
\begin{eqnarray}
\mathfrak{L}(\mathbf{x},\boldsymbol{\gamma}) = \log|\lambda \mathbf{I} + \mathbf{\Phi} \mathbf{\Sigma}_0 \mathbf{\Phi}^T  | + \frac{1}{\lambda} \|\mathbf{y}-\mathbf{\Phi}\mathbf{x}\|_2^2 + \mathbf{x}^T \mathbf{\Sigma}_0^{-1} \mathbf{x} . \nonumber
\end{eqnarray}
By first minimizing over $\boldsymbol{\gamma}$ and then minimizing over $\mathbf{x}$, we have:
\begin{eqnarray}
\mathbf{x} = \arg\min_\mathbf{x} \Big\{ \|\mathbf{y}-\mathbf{\Phi}\mathbf{x}\|_2^2 + \lambda g_{\mathrm{c}}(\mathbf{x}) \Big\},
\label{equ:x_space_expression}
\end{eqnarray}
with  the penalty $g_{\mathrm{c}}(\mathbf{x})$ given by
\begin{eqnarray}
g_{\mathrm{c}}(\mathbf{x}) \triangleq \min_{\boldsymbol{\gamma} \succeq \mathbf{0}}  \Big\{ \mathbf{x}^T \mathbf{\Sigma}_0^{-1} \mathbf{x}  + \log|\lambda \mathbf{I} + \mathbf{\Phi} \mathbf{\Sigma}_0 \mathbf{\Phi}^T  | \Big\}.
\label{equ:g_originalExp}
\end{eqnarray}
Define $h(\boldsymbol{\gamma}) \triangleq \log|\lambda \mathbf{I} + \mathbf{\Phi}\mathbf{\Sigma}_0 \mathbf{\Phi}^T|$. It is concave and non-decreasing w.r.t. $\boldsymbol{\gamma} \succeq \mathbf{0} $. Thus we have
\begin{eqnarray}
\log|\lambda \mathbf{I} + \mathbf{\Phi}\mathbf{\Sigma}_0 \mathbf{\Phi}^T| = \min_{\mathbf{z} \succeq \mathbf{0}} \mathbf{z}^T \boldsymbol{\gamma} - h^*(\mathbf{z})  \label{equ:dual_of_Sigmay}
\end{eqnarray}
where $h^*(\mathbf{z})$ is the concave conjugate of $h(\boldsymbol{\gamma})$ and can be expressed as $h^*(\mathbf{z}) = \min_{\boldsymbol{\gamma} \succeq \mathbf{0}} \mathbf{z}^T \boldsymbol{\gamma} - \log|\lambda \mathbf{I} + \mathbf{\Phi}\mathbf{\Sigma}_0 \mathbf{\Phi}^T|$. Thus, using (\ref{equ:dual_of_Sigmay})  we can express  (\ref{equ:g_originalExp}) as
\begin{eqnarray}
g_{\mathrm{c}}  (\mathbf{x})
&=& \min_{\boldsymbol{\gamma},\mathbf{z} \succeq \mathbf{0}}  \mathbf{x}^T \mathbf{\Sigma}_0^{-1} \mathbf{x} + \mathbf{z}^T \boldsymbol{\gamma} - h^*(\mathbf{z}) \nonumber \\
&=& \min_{\boldsymbol{\gamma},\mathbf{z} \succeq \mathbf{0}} \sum_{i}\Big(\frac{\mathbf{x}_i^T \mathbf{B}_i^{-1} \mathbf{x}_i}{\gamma_i} + z_i \gamma_i  \Big) - h^*(\mathbf{z}). \label{equ:g_final}
\end{eqnarray}
Minimizing (\ref{equ:g_final}) over $\gamma_i$, we have
\begin{eqnarray}
\gamma_i = z_i^{-\frac{1}{2}} \sqrt{\mathbf{x}_i^T \mathbf{B}_i^{-1} \mathbf{x}_i} \quad (\forall i)
\label{equ:gamma_i_by_zi}
\end{eqnarray}
Substituting (\ref{equ:gamma_i_by_zi}) into (\ref{equ:g_final}) leads to
\begin{eqnarray}
g_{\mathrm{c}}(\mathbf{x}) = \min_{\mathbf{z} \succeq \mathbf{0}} \sum_i \big( 2 z_i^{\frac{1}{2}} \sqrt{ \mathbf{x}_{i}^T \mathbf{B}_i^{-1} \mathbf{x}_i }  \big) - h^*(\mathbf{z}).
\label{equ:g_of_zi}
\end{eqnarray}
Using (\ref{equ:g_of_zi}), the problem (\ref{equ:x_space_expression}) now becomes:
\begin{eqnarray}
\mathbf{x} &=& \arg\min_\mathbf{x} \|\mathbf{y}-\mathbf{\Phi}\mathbf{x} \|_2^2  \nonumber\\
 && + \lambda \Big[ \min_{\mathbf{z} \succeq \mathbf{0}} \sum_i \big( 2 z_i^{\frac{1}{2}} \sqrt{\mathbf{x}_{i}^T \mathbf{B}_i^{-1} \mathbf{x}_i }  \big)- h^*(\mathbf{z})  \Big]. \label{equ:iterativeGroupLasso}
\end{eqnarray}

To further simplify the expression, we now calculate the optimal value of $z_i^{\frac{1}{2}}$. However, we need not calculate this value from the above expression. According to the duality property, from the relation (\ref{equ:dual_of_Sigmay}) we can directly obtain the  optimal value as follows:
\begin{eqnarray}
z_i^{\frac{1}{2}} &=& \Big( \frac{\partial \log| \lambda \mathbf{I} + \mathbf{\Phi}\mathbf{\Sigma}_0 \mathbf{\Phi}^T|}{\partial \gamma_i} \Big)^{\frac{1}{2}}  \nonumber \\
 &=&  \Big( \mathrm{Tr}\big[\mathbf{B}_i {\mathbf{\Phi}^i}^T \big( \lambda \mathbf{I} + \mathbf{\Phi}\mathbf{\Sigma}_0 \mathbf{\Phi}^T)^{-1} \mathbf{\Phi}^i \big] \Big)^{\frac{1}{2}}.
\label{equ:z_uprule}
\end{eqnarray}
Note that $z_i$ is a function of $\boldsymbol{\gamma}$, while according to (\ref{equ:gamma_i_by_zi}) $\gamma_i$ is a function of $\mathbf{x}_i$ (and $z_i$). This means that the problem (\ref{equ:iterativeGroupLasso}) should be solved in an iterative way. In the $k$-th iteration,  having used the update rules (\ref{equ:gamma_i_by_zi}) and (\ref{equ:z_uprule}) to obtain $(z_i^{(k)})^{1/2}$,  we need to solve the following optimization problem:
\begin{eqnarray}
\mathbf{x}^{(k+1)} = \arg\min_\mathbf{x} \, \|\mathbf{y}-\mathbf{\Phi}\mathbf{x}\|_2^2  + \lambda \sum_i  w_i^{(k)} \sqrt{ \mathbf{x}_i^T \mathbf{B}_i^{-1} \mathbf{x}_i }, \label{equ:SBL_L1version}
\end{eqnarray}
where $w_i^{(k)} \triangleq 2 (z_i^{(k)})^{1/2}$. And the resulting $\mathbf{x}^{(k+1)}$ will be used to update $\gamma_i$ and $z_i$, which are in turn used to calculate the solution in the next iteration.

The solution to (\ref{equ:SBL_L1version}) can be calculated using any Group-Lasso type algorithm. To see this, let $\mathbf{u}_i \triangleq w_i^{(k)} \mathbf{B}_i^{-1/2} \mathbf{x}_i$, $\mathbf{u} \triangleq [\mathbf{u}_1^T,\cdots,\mathbf{u}_g^T]^T$ and $\mathbf{H} \triangleq \mathbf{\Phi} \cdot \mathrm{diag}\{ \mathbf{B}_1^{1/2}/w_1^{(k)},\cdots, \mathbf{B}_g^{1/2}/w_g^{(k)} \}$. Then the problem (\ref{equ:SBL_L1version}) can be transformed to the following one:
\begin{eqnarray}
\mathbf{u}^{(k+1)}  = \arg\min_\mathbf{u} \| \mathbf{y} - \mathbf{H}\mathbf{u} \|_2^2 + \lambda \sum_i \|  \mathbf{u}_i \|_2. \label{equ:BSBL_L1_groupLasso_form}
\end{eqnarray}
Now each iteration is a standard Group-Lasso type problem, while the whole algorithm is an iterative reweighted algorithm.

In the above development we did not consider the learning rules for the regularizers $\lambda$ and $\mathbf{B}_i$. In fact, their estimation greatly benefits from this iterative reweighted form. Since each iteration is a Group-Lasso type problem, the optimal value of $\lambda$ can be automatically selected in the Group Lasso framework \cite{Tibshirani2012}. Also, since each iteration provides a block sparse solution, which is close to the true solution, $\mathbf{B}_i$ can be directly estimated from the solution of the previous iteration. In particular, each nonzero block in the previous solution can be treated as an AR(1) process, and its AR coefficient is thus estimated. The AR coefficients associated with all the nonzero blocks are averaged \footnote{The averaging is important. Otherwise, the algorithm may have poor performance.}, and the average value, denoted by $\bar{r}$, is used to construct each $\widehat{\mathbf{B}}_i$ according to (\ref{equ:EM_Brule_diffsize}).

The above algorithm is denoted by \textbf{BSBL-$\ell_1$}. It can be seen as a hybrid of a BSBL algorithm and a Group-Lasso type algorithm. On the one hand, it has the ability to adaptively learn and exploit intra-block correlation for better performance, as BSBL-EM and BSBL-BO. On the other hand, since it only takes few iterations (generally about 2 to 5 iterations in noisy environments) and each iteration can be implemented by any efficient Group-Lasso type algorithm, it is much faster and is more suitable for large-scale datasets than  BSBL-EM and BSBL-BO.

The algorithm also provides insights if we want to equip Group-Lasso type algorithms with the ability to exploit  intra-block correlation for better recovery performance.  We can consider this iterative reweighted method and change the $\ell_2$ norm of $\mathbf{x}_i$, i.e., $\|\mathbf{x}_i\|_2$, to the Mahalanobis distance type measure $\sqrt{\mathbf{x}_i^T \mathbf{B}_i^{-1} \mathbf{x}_i}$.

\section{Algorithms When the Block Partition is Unknown}
\label{sec:type2}

Now we extend the BSBL framework to address the situation when the block partition is unknown. For the algorithm development,
we assume that all the blocks are of equal size $h$ and the nonzeros blocks are arbitrarily located.
Later we will see that the approximation of equal block size is not limiting.
Note that though the resulting algorithms are not very sensitive to the choice of $h$, algorithmic performance can be further improved if a suitable value of $h$ is selected. We will comment more on $h$ later.

Given the identical block size $h$, there are $p \triangleq N-h+1$ possible (overlapping) blocks in $\mathbf{x}$. The $i$-th block starts at the $i$-th element of $\mathbf{x}$ and ends at the $(i+h-1)$-th element. All the nonzero elements of $\mathbf{x}$ lie within a subset of these blocks. Similar to Section \ref{sec:type1}, for the $i$-th block, we assume it satisfies a multivariate Gaussian distribution with the mean given by $\mathbf{0}$ and the covariance matrix given by $\gamma_i \mathbf{B}_i$, where $\mathbf{B}_i \in \mathbb{R}^{h \times h}$. So the prior of $\mathbf{x}$ has the form: $p(\mathbf{x}) \sim  \mathcal{N}_x(\textbf{0},\mathbf{\Sigma}_0)$. Note that due to the overlapping locations of these blocks, $\mathbf{\Sigma}_0$ is no longer a block diagonal matrix. It has the structure that each $\gamma_i \mathbf{B}_i$ lies along the principal diagonal of $\mathbf{\Sigma}_0$ and overlaps other neighboring $\gamma_j \mathbf{B}_j (j\neq i)$. Thus, we cannot directly use the BSBL framework and need to make some modifications.

To facilitate the use of the BSBL framework, we expand the covariance matrix $\mathbf{\Sigma}_0$ as follows:
\begin{eqnarray}
\widetilde{\mathbf{\Sigma}}_0 = \mathrm{diag}\{\gamma_1 \mathbf{B}_1,\cdots, \gamma_p \mathbf{B}_p\} \in \mathbb{R}^{ph \times ph}
\label{equ:newCov0}
\end{eqnarray}
Note that $\gamma_i \mathbf{B}_i$ no longer overlaps other $\gamma_j \mathbf{B}_j (i\neq j)$.
The  definition of $\widetilde{\mathbf{\Sigma}}_0$ implies the following decomposition of $\mathbf{x}$:
\begin{eqnarray}
\mathbf{x} = \sum_{i=1}^p \mathbf{E}_i \mathbf{z}_i,
\label{equ:newX}
\end{eqnarray}
where $\mathbf{z}_i \in \mathbb{R}^{h \times 1}$, $E\{ \mathbf{z}_i \} = \mathbf{0}$, $E\{ \mathbf{z}_i \mathbf{z}_j^T \} = \delta_{i,j} \gamma_i \mathbf{B}_i$ ($\delta_{i,j} = 1$ if $i=j$; otherwise, $\delta_{i,j} = 0$), and $\mathbf{z}\triangleq [\mathbf{z}_1^T,\cdots,\mathbf{z}_p^T]^T \sim  \mathcal{N}_z(\textbf{0},\widetilde{\mathbf{\Sigma}}_0)$. $\mathbf{E}_i \in \mathbb{R}^{N \times h}$ is a zero matrix except that the part from its $i$-th row to $(i+h-1)$-th row is replaced by the identity matrix $\mathbf{I}$. Then the original model (\ref{equ:SMV model}) can be expressed as:
\begin{eqnarray}
\mathbf{y} = \sum_{i=1}^p \mathbf{\Phi} \mathbf{E}_i \mathbf{z}_i + \mathbf{v}
\triangleq \mathbf{Az} + \mathbf{v},
\label{equ:new model}
\end{eqnarray}
where $\mathbf{A} \triangleq [\mathbf{A}_1,\cdots,\mathbf{A}_p]$ with $\mathbf{A}_i \triangleq \mathbf{\Phi} \mathbf{E}_i$. Now the new model (\ref{equ:new model}) is a block sparse model and can be solved by the BSBL framework. Thus,  following the development of BSBL-EM, BSBL-BO, and BSBL-$\ell_1$,  we obtain algorithms for this expanded model, which are called \textbf{EBSBL-EM}, \textbf{EBSBL-BO}, and \textbf{EBSBL-$\ell_1$}, respectively.

In the derivation above we assume that all blocks have the equal known size, $h$. However, this assumption is not crucial for practical use. When the size of a nonzero block of $\mathbf{x}$, say $\mathbf{x}_j$, is greater than or equal to $h$, it can be  recovered by a set of (overlapped) $\mathbf{z}_i$ ($i \in \mathcal{S}$, $\mathcal{S}$ is a non-empty set). When the size of $\mathbf{x}_j$ is less than $h$, it can be recovered by a $\mathbf{z}_i$ for some $i$. The experiments in Section \ref{sec:experiment} and in \cite{Zhilin_ICASSP2012} show that different values of $h$ lead to  similar performance.

The above insight also implies that even if the block partition is unknown, one can partition a signal into a number of non-overlapping blocks with user-defined block sizes, and then perform the BSBL algorithms. Nonetheless, performance of the BSBL algorithms are generally more sensitive to the block sizes than the EBSBL algorithms when recovering block sparse signals \cite{ZhilinThesis} \footnote{When directly recovering non-sparse signals, performance of the BSBL algorithms is not sensitive to block sizes \cite{zhang_ECG}.}.

Use of the expanded model when the block partition is unknown is quite different from existing approaches \cite{bcs_mcmc,Huang2009,faktor2010exploiting}. Our new approach has several advantages. Firstly,  it simplifies the algorithms, which, in turn, increases robustness in noisy environments, as shown in Section \ref{sec:experiment}. Secondly, it facilitates exploitation of intra-block correlation. Intra-block correlation is common in practical applications. Exploiting such correlation can significantly improve performance, yielding an advantage to our approach over existing methods which ignore intra-block correlation.

\begin{figure}[tbp]
\begin{minipage}[b]{0.48\linewidth}
  \centering
  \centerline{\epsfig{figure=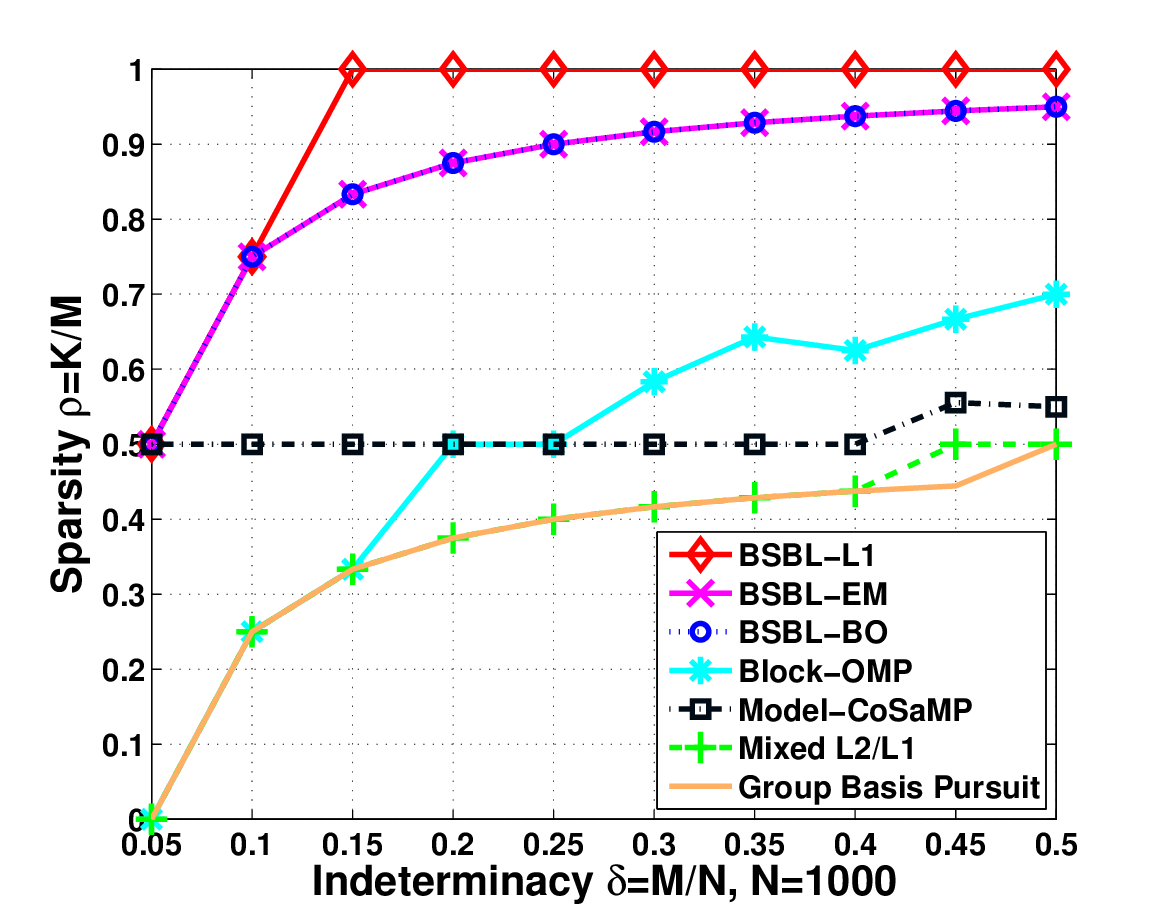,width=4.8cm}}
  \centerline{\footnotesize{(a) Intra-Block Correlation: 0}}
\end{minipage}
\hfill
\begin{minipage}[b]{.48\linewidth}
  \centering
  \centerline{\epsfig{figure=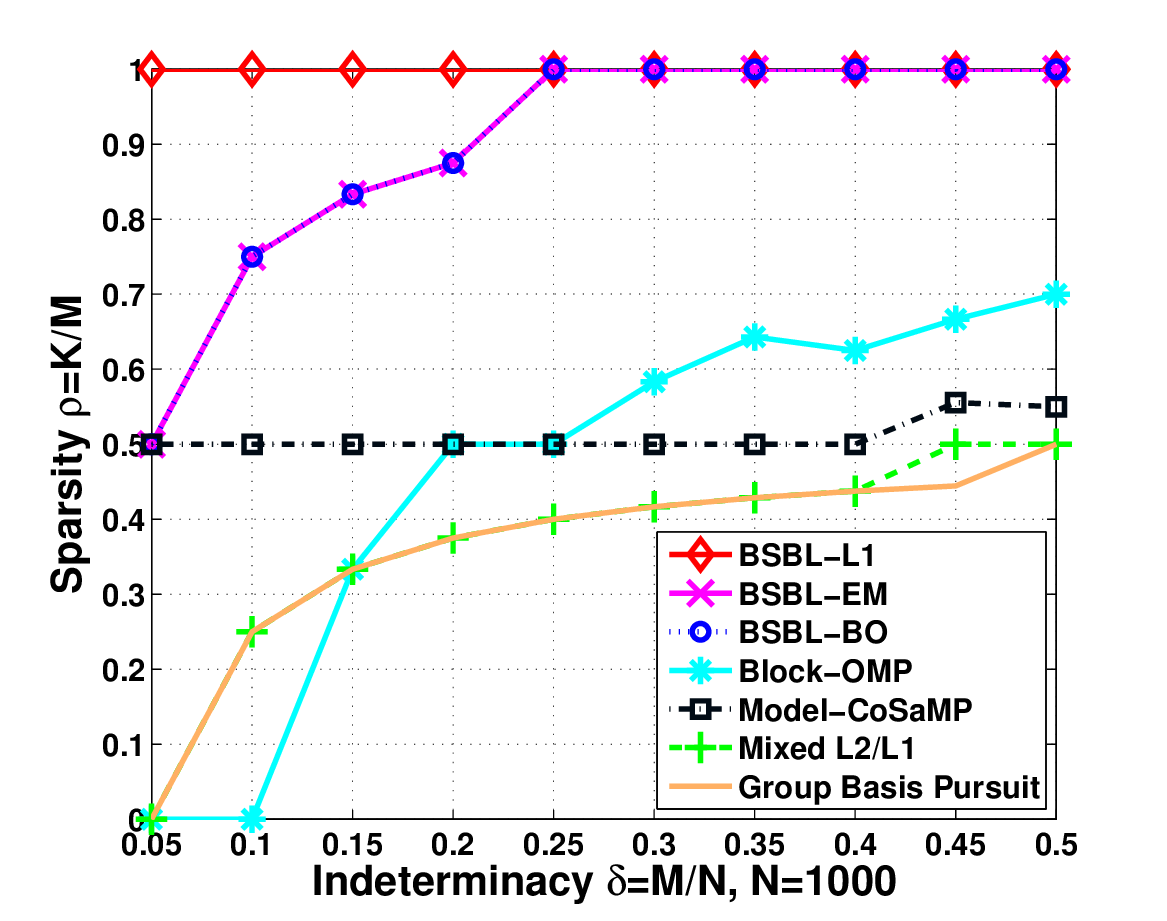,width=4.8cm}}
  \centerline{\footnotesize{(b) Intra-Block Correlation: 0.95}}
\end{minipage}
\caption{Empirical $99\%$ phase transitions of all the algorithms (a) when the intra-block correlation was 0, and (b) when the intra-block correlation was 0.95. Each point on a phase transition curve corresponds to the success rate larger than or equal to 0.99.}
\label{fig:phase}
\end{figure}

\section{Experiments}
\label{sec:experiment}

Due to space limitations, we only present some representative experimental results based on computer simulations \footnote{Matlab codes can be downloaded at \url{http://dsp.ucsd.edu/~zhilin/BSBL.html}.}. Experiments on real-world data can be found in \cite{zhang_ECG}.

In the following, each experiment was repeated for 400 trials. In each trial the matrix $\mathbf{\Phi}$ was generated as a zero mean random Gaussian matrix with columns normalized to unit $\ell_2$ norm. In noisy experiments the Normalized Mean Square Error (NMSE) was used as a performance index, defined by $\| \widehat{\mathbf{x}} - \mathbf{x}_{\mathrm{gen}}  \|_2^2/ \| \mathbf{x}_{\mathrm{gen}} \|_2^2$, where $\widehat{\mathbf{x}}$ was the estimate of the true signal $\mathbf{x}_{\mathrm{gen}}$. In noiseless experiments the success rate was used as a performance index, defined as the percentage of successful trials in the 400 trials (A successful trial was defined as the one when $\mathrm{NMSE} \leq 10^{-5}$).

In noiseless experiments, we chose Mixed $\ell_2/\ell_1$ Program \cite{Eldar2009} to solve (\ref{equ:BSBL_L1_groupLasso_form}) in each iteration of BSBL-$\ell_1$; in noisy experiments, we chose Group Basis Pursuit for this purpose. For all of our algorithms, when calculating $r$, instead of using the original formula $r = \frac{m_1}{m_0}$, the formula $r \triangleq \mathrm{sign}(\frac{m_1}{m_0}) \min\{|\frac{m_1}{m_0}|,0.99\}$ was used to ensure that the calculated $r$ satisfies $-1<r<1$. The same modification applies to $\bar{r}$.

\subsection{Phase Transition}
\label{subsec:phase}

We first examined empirical phase transitions \cite{donoho2009observed} \footnote{The phase transition graph was initially used to describe each algorithm's ability to recover a sparse signal with no structure. In this experiment it was used to describe each algorithm's ability to recover a block sparse signal.} in exact recovery of  block sparse signals in noiseless environments for our three BSBL algorithms, Block-OMP, Model-CoSaMP, Mixed $\ell_2/\ell_1$ Program, and Group Basis Pursuit. The phase transition is used to illustrate how sparsity level (defined as $\rho=K/M$, where $K$ is the number of nonzero elements in $\mathbf{x}$) and indeterminacy (defined as $\delta=M/N$) affect each algorithm's success in the exact recovery. Each point on the plotted phase transition curve corresponds to an algorithm's success rate greater than or equal to $99\%$ in 400 trials. Above the curve the success rate sharply drops.

In the experiment we varied the indeterminacy $\delta=M/N$ from 0.05 to 0.5 with $N$ fixed to 1000. For each $M$ and $N$, a block sparse signal was generated, which consisted of 40 blocks with an identical block size of 25 elements. The number of nonzero blocks varied from 1 to 20; thus the number of nonzero elements varied from 25 to 500. The locations of the nonzero blocks were determined randomly. The block partition was known to the algorithms, but the number of nonzero blocks and their locations were unknown to the algorithms. Each nonzero block satisfied a multivariate Gaussian distribution with zero mean and covariance matrix $\mathbf{\Sigma}_{\mathrm{gen}}$. By manipulating the covariance matrix, and thus changing intra-block correlation, we examined the effect of intra-block correlation on each algorithm's phase transition.

We first considered the situation when the intra-block correlation was 0 (i.e., $\mathbf{\Sigma}_{\mathrm{gen}}=\mathbf{I}$). The empirical phase transition curves of all the algorithms are shown in Fig.\ref{fig:phase} (a). We can see that the three BSBL algorithms had the best performance, and the phase transition curves of BSBL-EM and BSBL-BO were identical. It is worth noting that when $\delta \geq 0.15$, BSBL-$\ell_1$ exactly recovered block sparse signals with $\rho=1$ with a high success rate ($\geq 99\%$).

The results become more interesting when the intra-block correlation was 0.95 (i.e., $\mathbf{\Sigma}_{\mathrm{gen}}=\mathrm{Toeplitz}([1,0.95,\cdots,0.95^{24}])$). The empirical phase transition curves are shown in Fig.\ref{fig:phase} (b), where all the three BSBL algorithms had improved performance. BSBL-$\ell_1$  exactly recovered sparse signals with $\rho=1$ even for $\delta<0.15$. BSBL-EM and BSBL-BO could exactly recover sparse signals with $\rho=1$ when $\delta \geq 0.25$. In contrast, all the four non-BSBL algorithms showed little change in performance when the intra-block correlation changed from 0 to 0.95.

These results are very interesting and surprising, since this may be the first time that an algorithm shows the ability to recover a block sparse signal of $M$ nonzero elements from $M$ measurements with a high success rate ($\geq 99\%$). Obviously, exploiting the block structure and the intra-block correlation plays a crucial role here, indicating the advantages of the BSBL framework.

\begin{figure}[tbp]
\begin{minipage}[b]{.48\linewidth}
  \centering
  \centerline{\epsfig{figure=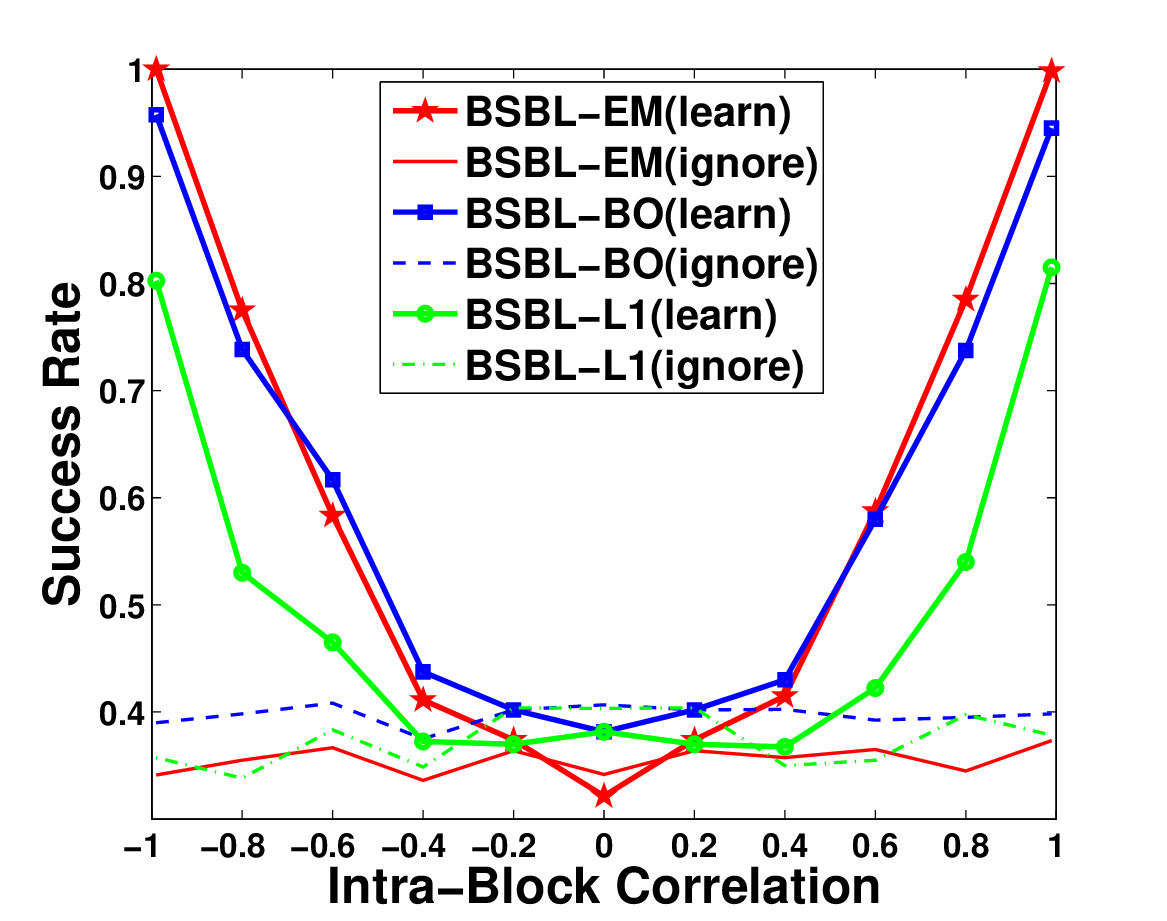,width=4.8cm}}
  \centerline{\footnotesize{(a)}}
\end{minipage}
\hfill
\begin{minipage}[b]{0.48\linewidth}
  \centering
  \centerline{\epsfig{figure=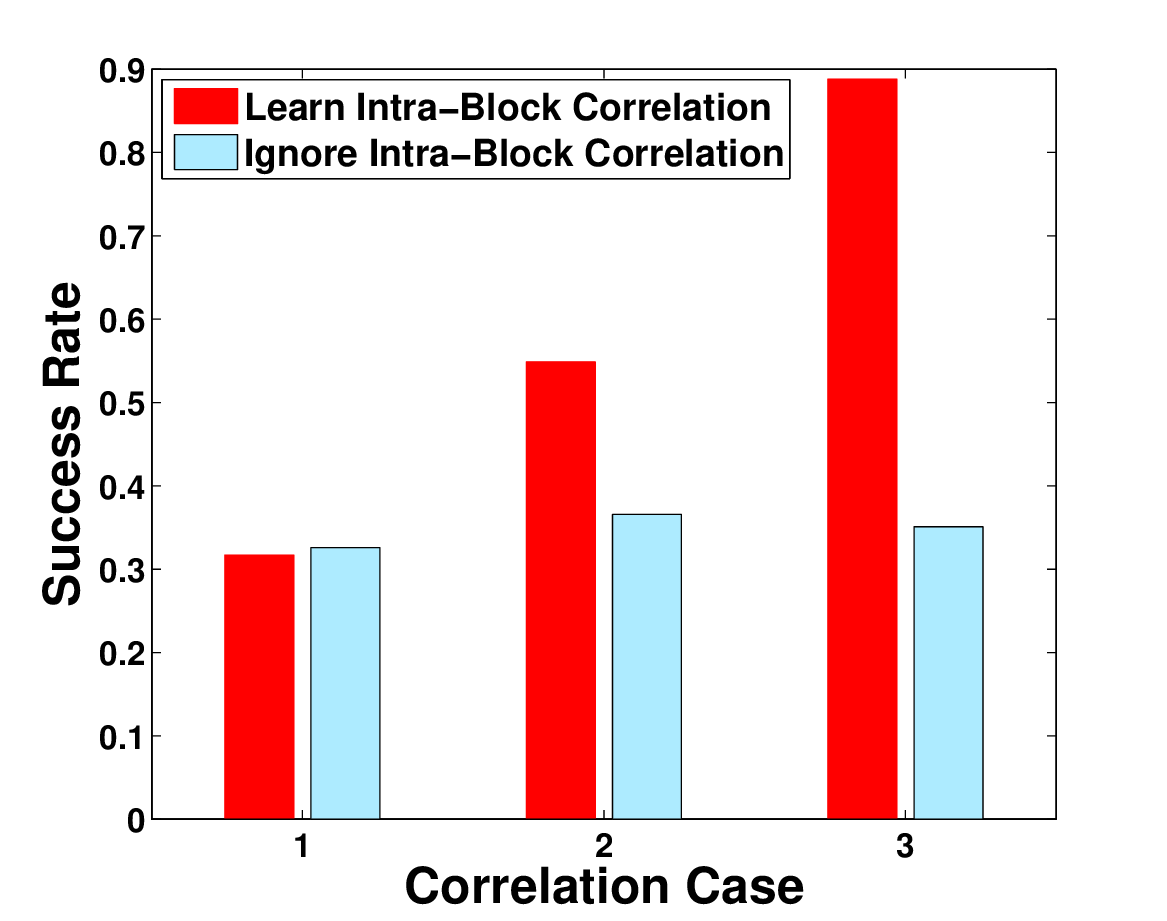,width=4.8cm}}
  \centerline{\footnotesize{(b)}}
\end{minipage}
\caption{(a) shows the benefit of exploiting the intra-block correlation. (b) shows the performance of BSBL-EM for three correlation conditions.}
\label{fig:corr}
\end{figure}

\subsection{Benefit of Exploiting Intra-Block Correlation}

The above results suggest there is a benefit to exploiting intra-block correlation. To further clarify this, another noiseless experiment was carried out. The matrix $\mathbf{\Phi}$ was  of the size $100 \times 300$. The signal consisted of 75 blocks with an identical size of 4 elements. Only 20 of the blocks were nonzero. All the nonzero blocks had the same intra-block correlation (generated as in Section \ref{subsec:phase}) ranging from -0.99 to 0.99. Different from the first experiment, each nonzero block was further normalized to unit $\ell_2$ norm in order to remove the interference caused by different $\ell_2$ norms of the blocks.

BSBL-EM, BSBL-BO and BSBL-$\ell_1$ were applied with and without correlation exploitation. In the first case, they adaptively learned and exploited the intra-block correlation. In the second case, they ignored the correlation, i.e., fixing $\mathbf{B}_i = \mathbf{I} (\forall i)$.

The results are shown in Fig.\ref{fig:corr} (a). First, we see that exploiting the intra-block correlation greatly improved the performance of the BSBL algorithms. Second, when ignoring the intra-block correlation, the performance of the BSBL algorithms showed no obvious relation to the correlation \footnote{This phenomenon can also be observed from the performance of the compared algorithms in Section \ref{subsec:phase}, where their performance had little change when intra-block correlation dramatically varied.}. In other words, no obvious negative effect is observed if ignoring the intra-block correlation. Note that the second observation is quite different from the observation on \emph{temporal correlation} in an MMV model \cite{Zhilin2011c}, where we found that if temporal correlation is not exploited, algorithms have poorer performance with increasing temporal correlation values \footnote{Temporal correlation in an MMV model can be viewed as intra-block correlation in a vectorized MMV model (which is a block sparse model). However, it should be noted that the basis matrix in the vectorized MMV model has the specific structure $\mathbf{\Phi} \otimes \mathbf{I}_L$ \cite{Zhilin2011c}, where $\mathbf{\Phi}$ is the basis matrix in the original MMV model, $\otimes$ indicates the Kronecker product, $\mathbf{I}_L$ is the identity matrix with the dimension $L\times L$, and $L$ is the number of measurement vectors in the MMV model. This structure is not present in the block sparse model considered in this work, which is believed to account for the different behavior with respect to the intra-block correlation investigated here.}.

In the previous experiment all the generated nonzero blocks had the same intra-block correlation. We might then ask whether the proposed algorithms can still succeed when the intra-block correlation for nonzero blocks is not homogenous. To answer this question, we considered three cases for generating each nonzero block: (1) the intra-block correlation values were chosen uniformly randomly from -1 to 1; (2) the correlation values were chosen uniformly randomly from 0 to 1; (3) the correlation values were chosen uniformly randomly from 0.7 to 1.

BSBL-EM was then applied with and without correlation exploitation, as described in the previous experiment. The results are shown in Fig.\ref{fig:corr} (b), with the three correlation cases indicated by `Case 1', `Case 2', and `Case 3', respectively. We can see in Case 3 (least variation in intra-block correlation values) the benefit of exploiting the correlation was significant, while in Case 1 (most variation in intra-block correlation values) the benefit disappeared, but exploiting the correlation was not harmful. However, Case 1 rarely happens in practice. In most practical problems the intra-block correlation values of all nonzero blocks tends to be positive and high, which corresponds to Case 2 and Case 3.

\subsection{Performance in Noisy Environments}

We compared the BSBL algorithms, Mixed $\ell_2/\ell_1$ Program, Group Lasso, and Group Basis Pursuit at different noise levels. In this experiment $M=128$ and $N=512$.  The generated block sparse signal was partitioned into 64 blocks with an identical block size of 8 elements. Seven blocks were nonzero, generated as in Section \ref{subsec:phase}. The intra-block correlation value for each block was uniformly randomly varied from 0.8 to 1. Gaussian white noise was added such that the SNR, defined by $\mathrm{SNR}(\mathrm{dB}) \triangleq 20\log_{10} (\|\mathbf{\Phi} \mathbf{x}_{\mathrm{gen}}\|_2 / \| \mathbf{v} \|_2 )$, ranged from 5 dB to 25 dB for each generated signal. As a benchmark result, the `oracle' result was calculated, which was the least-square estimate of $\mathbf{x}_{\mathrm{gen}}$ given its true support.

\begin{figure}[tbp]
\begin{minipage}[b]{.48\linewidth}
  \centering
  \centerline{\epsfig{figure=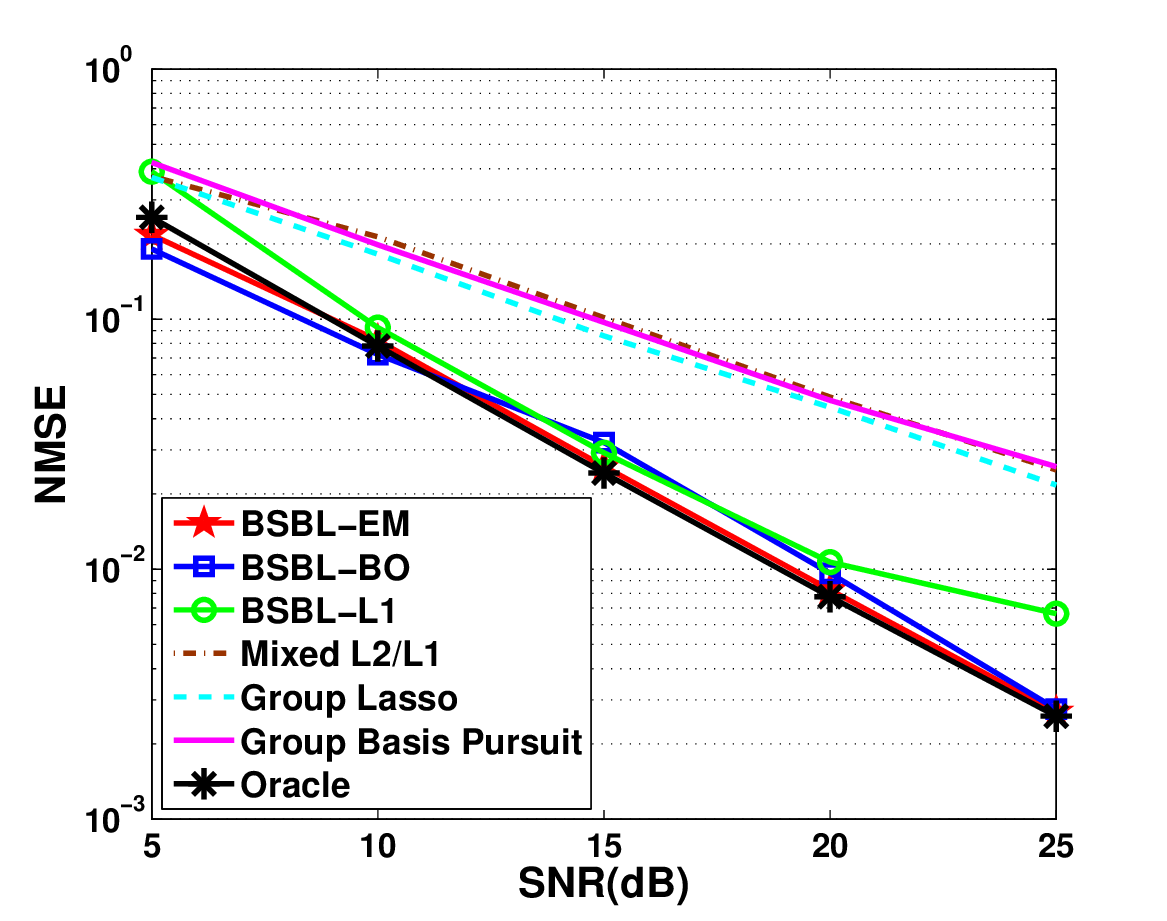,width=4.8cm}}
  \centerline{\footnotesize{(a)}}
\end{minipage}
\hfill
\begin{minipage}[b]{0.48\linewidth}
  \centering
  \centerline{\epsfig{figure=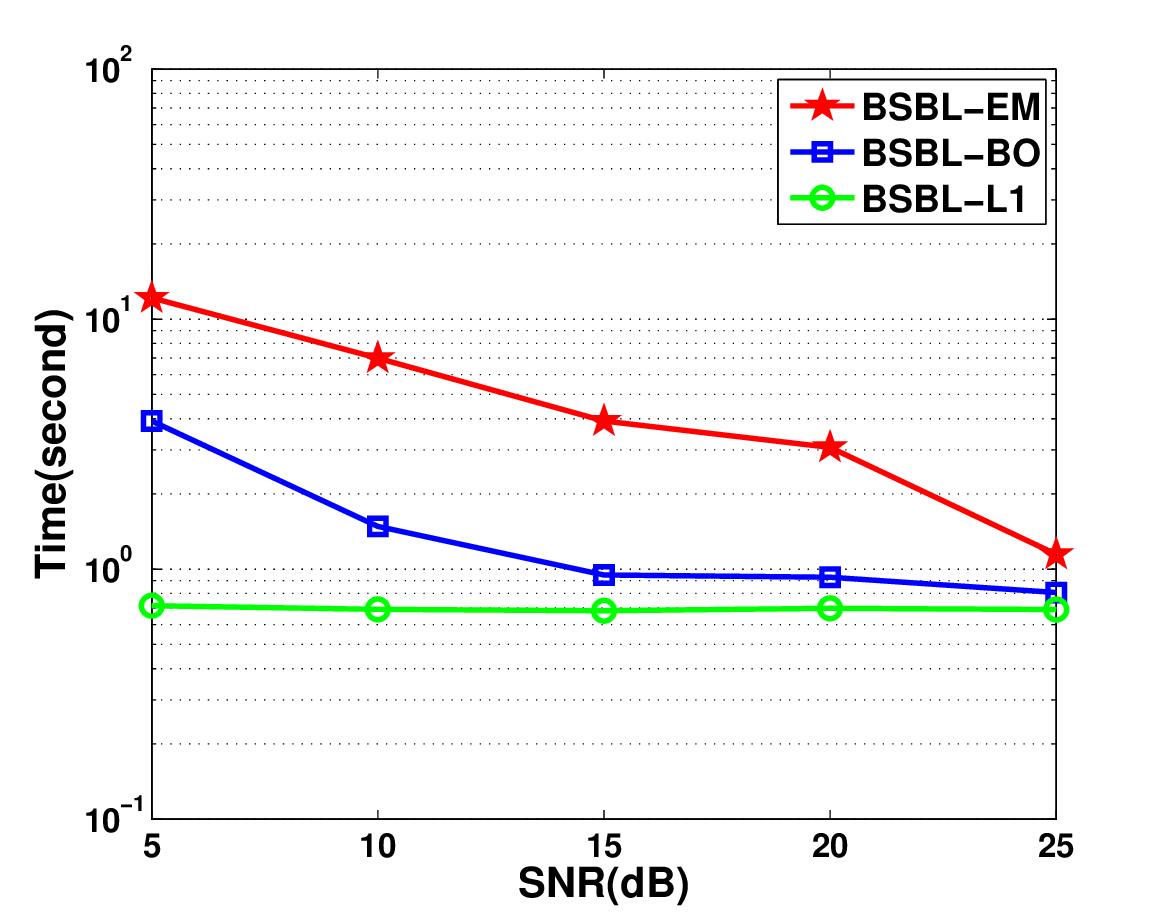,width=4.8cm}}
  \centerline{\footnotesize{(b)}}
\end{minipage}
\caption{(a) Performance comparison in different noise levels. (b) Comparison of the computational speed  of the three BSBL algorithms in the noisy experiment.}
\label{fig:noisy}
\end{figure}

The results are shown in Fig.\ref{fig:noisy} (a). All three BSBL algorithms exhibited significant performance gains over non-BSBL algorithms. In particular, the performance curves of BSBL-EM and BSBL-BO were nearly identical to that of the 'oracle'. The phenomenon that BSBL-$\ell_1$ had slightly poorer performance at low SNR and high SNR situations is due to some sub-optimal default parameters in the software implementing Group Basis Pursuit \cite{van2008probing}. We found the phenomenon disappeared when using other software. Figure \ref{fig:noisy} (b) gives the speed comparison of the three algorithms on a computer with dual-core 2.8 GHz CPU, 6.0 GiB RAM, and Windows 7 OS. It shows BSBL-$\ell_1$ was the fastest due to the use of Group Basis Pursuit in its inner loop.

\begin{figure}[tbp]
\centering
\includegraphics[height=6cm,width=7.5cm]{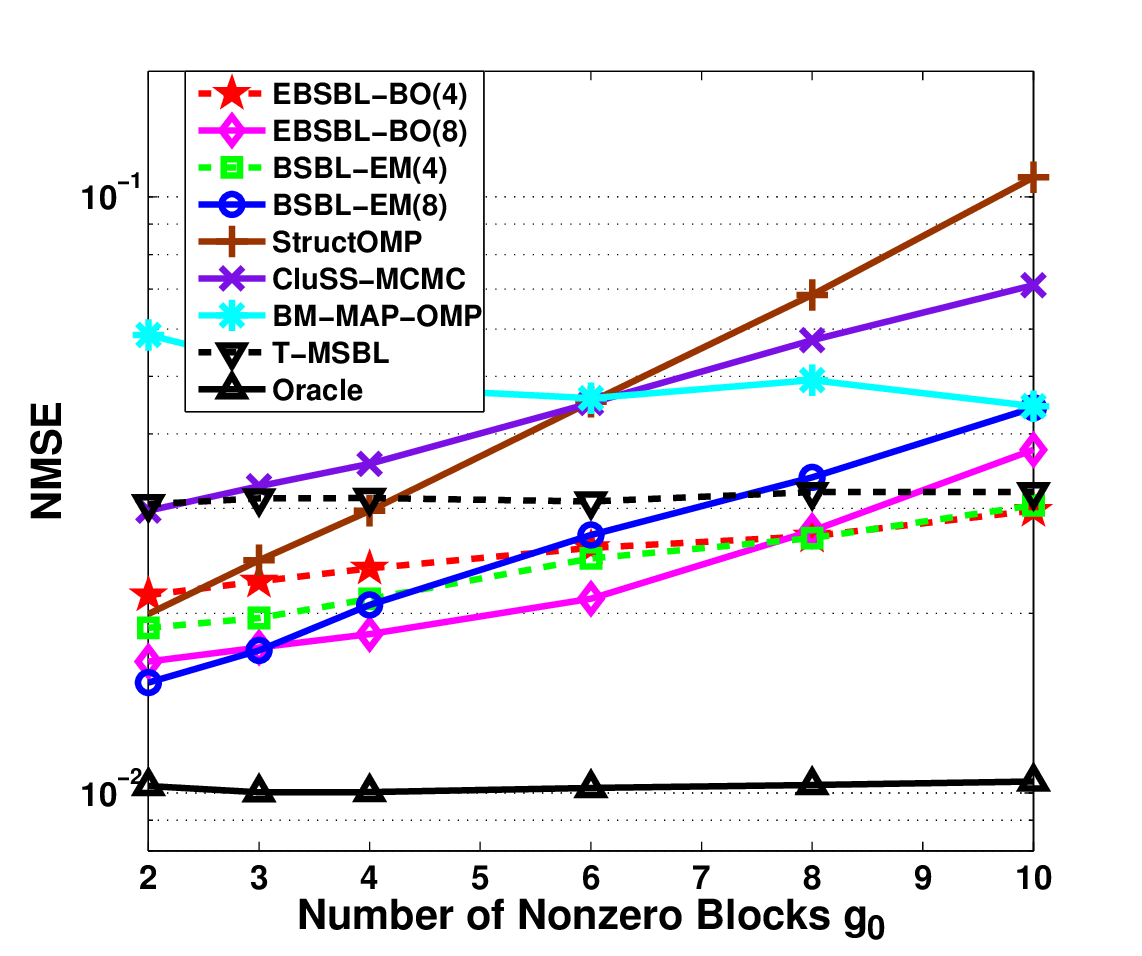}
\caption{Performance comparison when block partition was unknown.}
\label{fig:randomsize}
\end{figure}

\subsection{Performance When Block Partition Is Unknown}

We set up a noisy experiment to compared all of our algorithms with  StructOMP (given the number of nonzero elements), BM-MAP-OMP (given the true noise variance), and CluSS-MCMC, under the conditions where the block partition is unknown. The matrix $\mathbf{\Phi}$ was of the size $192 \times 512$. The signal $\mathbf{x}_{\mathrm{gen}}$ contained $g_0$ nonzero blocks with random size and random locations (not overlapping). $g_0$ was varied from 2 to 10. The total number of nonzero elements in $\mathbf{x}_{\mathrm{gen}}$ was fixed to 48. The intra-block correlation value for each block uniformly randomly varied from 0.8 to 1. SNR was 15 dB. As we stated in Section \ref{sec:type2}, knowledge of the block size $h$ is not crucial in practical use. To empirically evaluate this, we calculated performance curves for all our algorithms using fixed values of $h=4$ and $h=8$. The results are shown in Fig.\ref{fig:randomsize}. To improve figure readability, we only display BSBL-EM and EBSBL-BO. We also applied T-MSBL \cite{Zhilin2011c} here. Note that when T-MSBL is used for the block sparse signal recovery problem (\ref{equ:SMV model}), it can be viewed as a special case of BSBL-EM with $h=1$. The results show that our algorithms outperformed StructOMP, CluSS-MCMC, and BM-MAP-OMP. The results also show that for both BSBL-EM and EBSBL-BO, setting $h=4$ or $h=8$ led to similar performance.

\section{Conclusion}

Using the block sparse Bayesian learning framework and its extension, we proposed a number of algorithms to recover block sparse signals when the block structure is known or unknown. These  algorithms have the ability to explore and exploit intra-block correlation in signals and thereby improve performance. We experimentally demonstrated that these algorithms significantly outperform existing algorithms. The derived algorithms also suggest that the iterative reweighted framework  is a promising method for Group-Lasso type algorithms to exploit intra-block correlation.

\bibliographystyle{IEEEtran}

\bibliography{bibfile}

\begin{thebibliography}{10}
\providecommand{\url}[1]{#1}
\csname url@samestyle\endcsname
\providecommand{\newblock}{\relax}
\providecommand{\bibinfo}[2]{#2}
\providecommand{\BIBentrySTDinterwordspacing}{\spaceskip=0pt\relax}
\providecommand{\BIBentryALTinterwordstretchfactor}{4}
\providecommand{\BIBentryALTinterwordspacing}{\spaceskip=\fontdimen2\font plus
\BIBentryALTinterwordstretchfactor\fontdimen3\font minus
  \fontdimen4\font\relax}
\providecommand{\BIBforeignlanguage}[2]{{%
\expandafter\ifx\csname l@#1\endcsname\relax
\typeout{** WARNING: IEEEtran.bst: No hyphenation pattern has been}%
\typeout{** loaded for the language `#1'. Using the pattern for}%
\typeout{** the default language instead.}%
\else
\language=\csname l@#1\endcsname
\fi
#2}}
\providecommand{\BIBdecl}{\relax}
\BIBdecl

\bibitem{elad2010sparse}
M.~Elad, \emph{Sparse and redundant representations}.\hskip 1em plus 0.5em
  minus 0.4em\relax Springer Verlag, 2010.

\bibitem{grouplasso}
M.~Yuan and Y.~Lin, ``Model selection and estimation in regression with grouped
  variables,'' \emph{J. R. Statist. Soc. B}, vol.~68, pp. 49--67, 2006.

\bibitem{ModelCS}
R.~G. Baraniuk, V.~Cevher, M.~F. Duarte, and C.~Hegde, ``Model-based
  compressive sensing,'' \emph{IEEE Trans. on Information Theory}, vol.~56,
  no.~4, pp. 1982--2001, 2010.

\bibitem{BOMP}
Y.~C. Eldar, P.~Kuppinger, and H.~Bolcskei, ``Block-sparse signals: uncertainty
  relations and efficient recovery,'' \emph{IEEE Trans. on Signal Processing},
  vol.~58, no.~6, pp. 3042--3054, 2010.

\bibitem{van2008probing}
E.~Van Den~Berg and M.~Friedlander, ``Probing the pareto frontier for basis
  pursuit solutions,'' \emph{SIAM Journal on Scientific Computing}, vol.~31,
  no.~2, pp. 890--912, 2008.

\bibitem{Eldar2009}
Y.~C. Eldar and M.~Mishali, ``Robust recovery of signals from a structured
  union of subspaces,'' \emph{IEEE Trans. on Information Theory}, vol.~55,
  no.~11, pp. 5302--5316, 2009.

\bibitem{Huang2009}
J.~Huang, T.~Zhang, and D.~Metaxas, ``Learning with structured sparsity,'' in
  \emph{ICML 2009}, 2009, pp. 417--424.

\bibitem{bcs_mcmc}
L.~Yu, H.~Sun, J.~P. Barbot, and G.~Zheng, ``Bayesian compressive sensing for
  cluster structured sparse signals,'' \emph{Signal Processing}, vol.~92,
  no.~1, pp. 259--269, 2012.

\bibitem{faktor2010exploiting}
T.~Peleg, Y.~Eldar, and M.~Elad, ``Exploiting statistical dependencies in
  sparse representations for signal recovery,'' \emph{IEEE Trans. on Signal
  Processing}, vol.~60, no.~5, pp. 2286--2303, 2012.

\bibitem{zhang_ECG}
Z.~Zhang, T.-P. Jung, S.~Makeig, and B.~D. Rao, ``Compressed sensing for
  energy-efficient wireless telemonitoring of noninvasive fetal {ECG} via block
  sparse {Bayesian} learning,'' \emph{IEEE Trans. on Biomedical Engineering},
  vol.~60, no.~2, pp. 300--309, 2013.

\bibitem{Zhilin2011c}
Z.~Zhang and B.~D. Rao, ``Sparse signal recovery with temporally correlated
  source vectors using sparse {Bayesian} learning,'' \emph{IEEE Journal of
  Selected Topics in Signal Processing}, vol.~5, no.~5, pp. 912--926, 2011.

\bibitem{Cotter2005}
S.~F. Cotter, B.~D. Rao, K.~Engan, and K.~Kreutz-Delgado, ``Sparse solutions to
  linear inverse problems with multiple measurement vectors,'' \emph{IEEE
  Trans. on Signal Processing}, vol.~53, no.~7, pp. 2477--2488, 2005.

\bibitem{Zhilin_ICASSP2012}
Z.~Zhang and B.~D. Rao, ``Recovery of block sparse signals using the framework
  of block sparse {Bayesian} learning,'' in \emph{ICASSP 2012}, Japan, 2012,
  pp. 3345--3348.

\bibitem{Tipping2001}
M.~E. Tipping, ``Sparse {Bayesian} learning and the relevance vector machine,''
  \emph{J. of Mach. Learn. Res.}, vol.~1, pp. 211--244, 2001.

\bibitem{stoica2011spice}
P.~Stoica and P.~Babu, ``{SPICE and LIKES}: Two hyperparameter-free methods for
  sparse-parameter estimation,'' \emph{Signal Processing}, vol.~92, no.~7, pp.
  1580--1590, 2012.

\bibitem{Zhilin2011b}
\BIBentryALTinterwordspacing
Z.~Zhang and B.~D. Rao, ``Exploiting correlation in sparse signal recovery
  problems: Multiple measurement vectors, block sparsity, and time-varying
  sparsity,'' in \emph{ICML 2011 Workshop on Structured Sparsity: Learning and
  Inference}, 2011. [Online]. Available: \url{arXiv:1105.0725}
\BIBentrySTDinterwordspacing

\bibitem{David2010}
D.~Wipf and S.~Nagarajan, ``Iterative reweighted $\ell_1$ and $\ell_2$ methods
  for finding sparse solutions,'' \emph{IEEE Journal of Selected Topics in
  Signal Processing}, vol.~4, no.~2, pp. 317--329, 2010.

\bibitem{Tibshirani2012}
R.~Tibshirani, J.~Bien, J.~Friedman, and et~al, ``Strong rules for discarding
  predictors in lasso-type problems,'' \emph{J. R. Statist. Soc. B}, vol.~74,
  2012.

\bibitem{ZhilinThesis}
Z.~Zhang, ``Sparse signal recovery exploiting spatiotemporal correlation,''
  \emph{Ph.D. Dissertation, University of California, San Diego}, 2012.

\bibitem{donoho2009observed}
D.~Donoho and J.~Tanner, ``Observed universality of phase transitions in
  high-dimensional geometry, with implications for modern data analysis and
  signal processing,'' \emph{Philosophical Transactions of the Royal Society
  A}, vol. 367, no. 1906, pp. 4273--4293, 2009.

\end{thebibliography}

\end{document}